\documentclass[letterpaper, 10 pt, conference]{ieeeconf}
\usepackage{geometry}
\usepackage{graphicx} 
\usepackage{caption}
\usepackage{subcaption}
\usepackage{url}
\usepackage[utf8]{inputenc}
\usepackage[english]{babel}
\usepackage{csquotes}
\usepackage{float}
\usepackage{multirow}
\usepackage{booktabs}
\usepackage{xcolor}
\usepackage{booktabs}
\usepackage{tabularx}
\usepackage{lipsum}
\usepackage{flushend}
\usepackage{float}
\usepackage{makecell, tabularx}
\newcolumntype{L}{>{\raggedright\arraybackslash}X}
\newcolumntype{C}{>{\centering\arraybackslash}X}
\usepackage{siunitx}
\usepackage{booktabs,ragged2e}
\usepackage{textcomp}
\usepackage{tabularx}
\usepackage{algorithm}
\usepackage{algpseudocode}
\usepackage{algcompatible}

\usepackage[flushleft]{threeparttable}
\IEEEoverridecommandlockouts              
\usepackage[utf8]{inputenc}
\usepackage[T1]{fontenc}

\title{Leveraging Federated Learning for \\Automatic Detection of Clopidogrel Treatment Failures}

\author{%
  Samuel Kim and Min Sang Kim \\ \\
  Cipherome Inc., San Jose, CA, U.S.A. \\
  \small \texttt{\{sam.kim, min.kim\}@cipherome.com} \\
}

\begin{document}

\maketitle

\begin{abstract}
The effectiveness of clopidogrel, a widely used antiplatelet medication, varies significantly among individuals, necessitating the development of precise predictive models to optimize patient care. In this study, we leverage federated learning strategies to address clopidogrel treatment failure detection. Our research harnesses the collaborative power of multiple healthcare institutions, allowing them to jointly train machine learning models while safeguarding sensitive patient data. Utilizing the UK Biobank dataset, which encompasses a vast and diverse population, we partitioned the data based on geographic centers and evaluated the performance of federated learning. Our results show that while centralized training achieves higher Area Under the Curve (AUC) values and faster convergence, federated learning approaches can substantially narrow this performance gap. Our findings underscore the potential of federated learning in addressing clopidogrel treatment failure detection, offering a promising avenue for enhancing patient care through personalized treatment strategies while respecting data privacy. This study contributes to the growing body of research on federated learning in healthcare and lays the groundwork for secure and privacy-preserving predictive models for various medical conditions.
\end{abstract}

\section{Introduction}
The prediction of adverse reactions to clopidogrel, a widely prescribed antiplatelet medication, is a critical challenge in modern healthcare. Individual responses to clopidogrel can vary significantly, leading to varying degrees of treatment efficacy and the potential for adverse events. Traditional approaches to adverse reaction prediction typically involve centralized data analysis, which entails the aggregation of patient data from diverse sources into a single, consolidated repository~\cite{kim2023detection, choi2024data}. However, this centralized approach presents inherent challenges, including data privacy concerns, security risks, and regulatory complexities, particularly in the realm of healthcare data.

In recent years, federated learning has emerged as a transformative paradigm for addressing these issues. Federated learning enables collaborative model training across decentralized data sources while keeping raw data localized and secure. In the context of predicting adverse reactions to clopidogrel, federated learning offers a promising avenue to leverage insights from multiple healthcare institutions without compromising patient confidentiality.

While initially proposed for machine learning in mobile environments~\cite{mcmahanCommunicationEfficientLearningDeep2017}, federated learning has garnered increasing attention within the healthcare domain, primarily owing to its privacy-preserving capabilities~\cite{xuFederatedLearningHealthcare2021}. Researchers have been actively exploring its applications in addressing a wide array of medical conditions, including Alzheimer's disease~\cite{alimeerzaFairPrivacyPreservingAlzheimer2022} and COVID-19~\cite{haMultiSiteSplitLearning2021, dayanFederatedLearningPredicting2021}. Pati {\em et al.} conducted a substantial-scale federated learning study aimed at developing a tumor boundary detection model~\cite{patiFederatedLearningEnables2022}. They spanned 71 geographically distinct sites across six continents and demonstrated the superiority of their approach over conventional methods.

This paper focuses on a novel approach to detect clopidogrel treatment failures that harnesses the power of federated learning strategies. We present a federated learning framework designed to facilitate joint model training across diverse healthcare institutions, allowing them to pool their collective knowledge without exposing sensitive patient data. In particular, we use patients' medical history including diagnoses, procedures, and prescriptions. Our approach seeks to enhance the accuracy and generalizability of clopidogrel adverse reaction prediction models by aggregating insights from a multitude of sources.

\begin{figure*}[t!]
    \centering
    \includegraphics[width=1\textwidth]{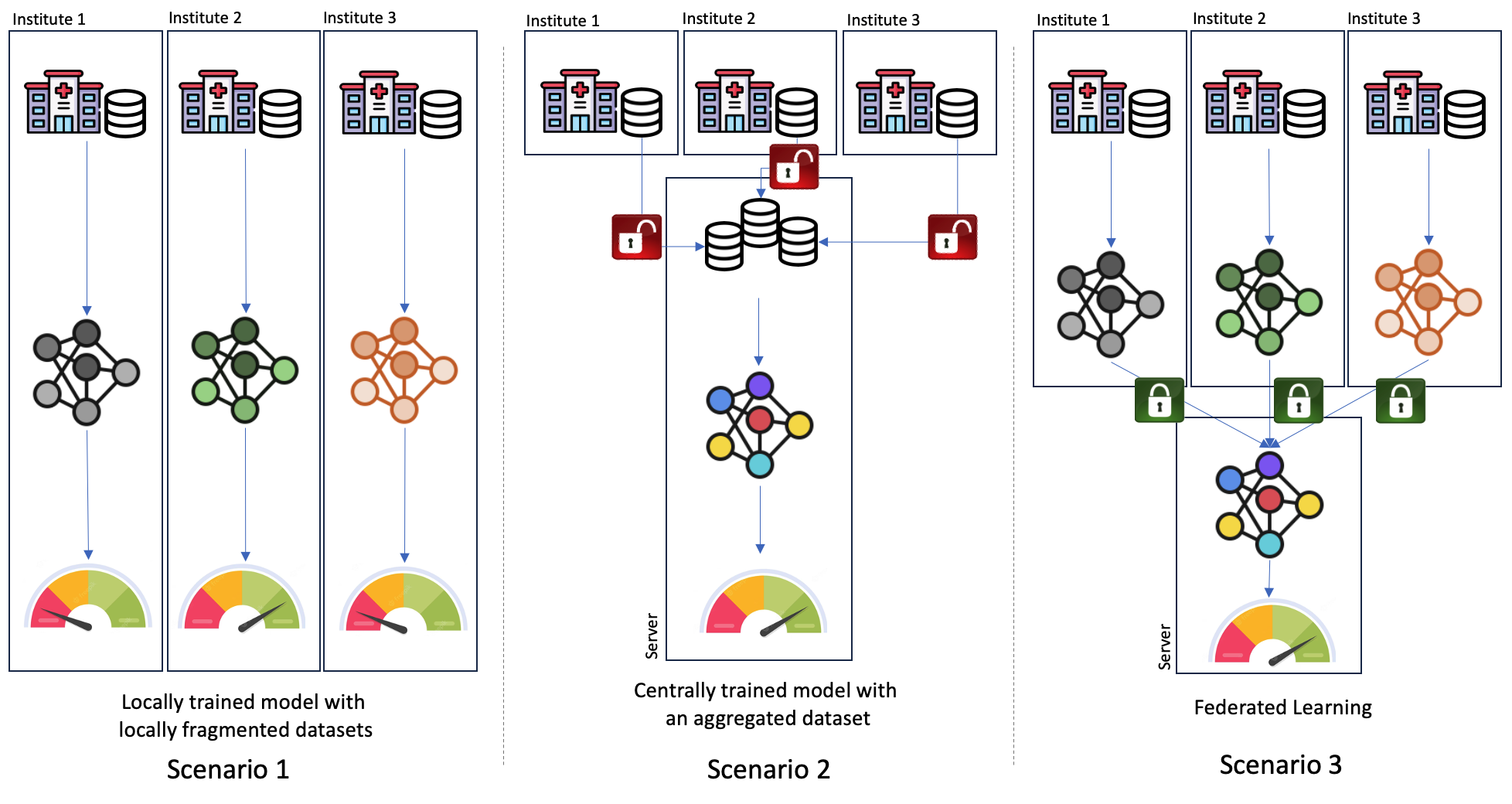}
    \caption{Experimental scenarios to illustrate how data is handled; 1) localized training, 2) centralized training, and 3) federated learning. }
    \label{fig:scenarios}
\end{figure*}

\section{Data}
\label{sec:data}
\subsection{Dataset}
We used the UK Biobank which consists of data collected from 502,527 participants ~\cite{UKB2015}. Volunteers aged 40 to 70 were recruited from England, Scotland, and Wales and invited to assessment centers between 2006 and 2010. Data collected during the visits included biosamples, physical examination measurements, and questionnaire answers followed by interviews. Genomic data, both sequencing and genotyping, were generated using the biosamples collected.  Also, an extensive and comprehensive medical history data was made available through hospital in-patient and primary care data from external data sources. 

We extracted prescriptions, diagnoses and procedure records, along with dates, for all participants from the UK Biobank. All prescription records were from general practitioner (GP) data coded in Read and British National Formulary (BNF) depending on the data supplier. Diagnoses and procedure records were from hospital in-patient records only and coded in ICD-9/10 and OPCS-3/4 respectively.

\subsection{Annotation} 
Treatment failure (TF) was defined as having a TF event within one year of the very first clopidogrel prescription. Clopidogrel prescriptions were identified using the substance or brand names of clopidogrel or respective read codes, while TF events include ischemic stroke, myocardial infarct, stent thrombosis and recurrent thrombosis or stenting. All of these annotations were carefully done by clinical professions reviewing patients' medical records. 

Subjects with events occurring within 7 days of the first prescription were excluded as it was unclear whether those events were associated with clopidogrel. The visit had to be through the emergency room to be valid in order to exclude follow up visits from previous events. From the dataset, we found 9,867 subjects with clopidogrel prescriptions. Among them, we labeled 1,824 patients as TF cases and 6,859 as control cases; 1,184 subjects were excluded due to data inconsistencies or ambiguities.  

\section{Methods}
\subsection{Experimental scenarios}

The study was conducted in three scenarios to show the benefits of the federated learning method as shown in Fig.~\ref{fig:scenarios}; 1) locally fragmented datasets, 2) an aggregated dataset as a whole, and 3) federated learning. The first and the second scenarios assume two opposite cases; the first is for the case that there is no way for researchers to share their datasets, while the second is for the case that researchers have full access to others’ datasets. We hypothesized that the performance in terms of accuracy between the two extreme scenarios is significantly different because of the lack of training data and model generalization that the first scenario inherently has. We also hypothesized that the performance of the federated learning method converges toward the one with the aggregated dataset as the proposed method should mitigate the problems with the fragmented datasets, therefore, attaining high predictive performance while ensuring the security of patient data.

Assuming that medical history data of individuals are stored in geographically distinct locations, we split the clopidogrel response dataset into 22 groups based on the UK Biobank assessment centre (field id 54-0.0) which includes the center location information of the initial assessment visit at which participants were recruited and consent given. The reason we chose the location of the first baseline survey is that most of participants only have the first baseline survey.

\begin{figure}[t!]
    \centering
    \includegraphics[width=\linewidth]{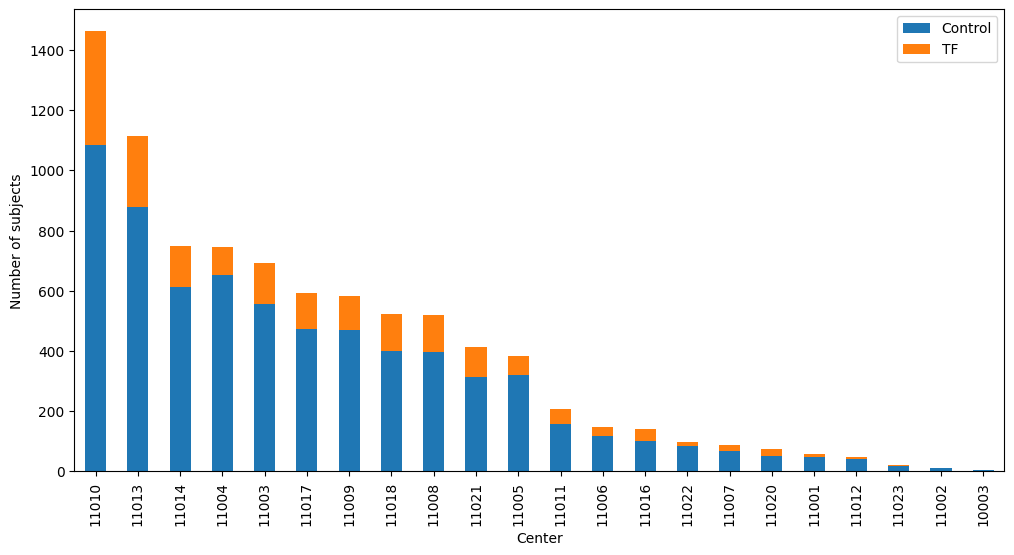}
    \caption{Number of patients with clopidogrel prescription in individual centers}
    \label{fig:counts}
\end{figure}

\begin{figure}[t]
    \includegraphics[width=1\linewidth]{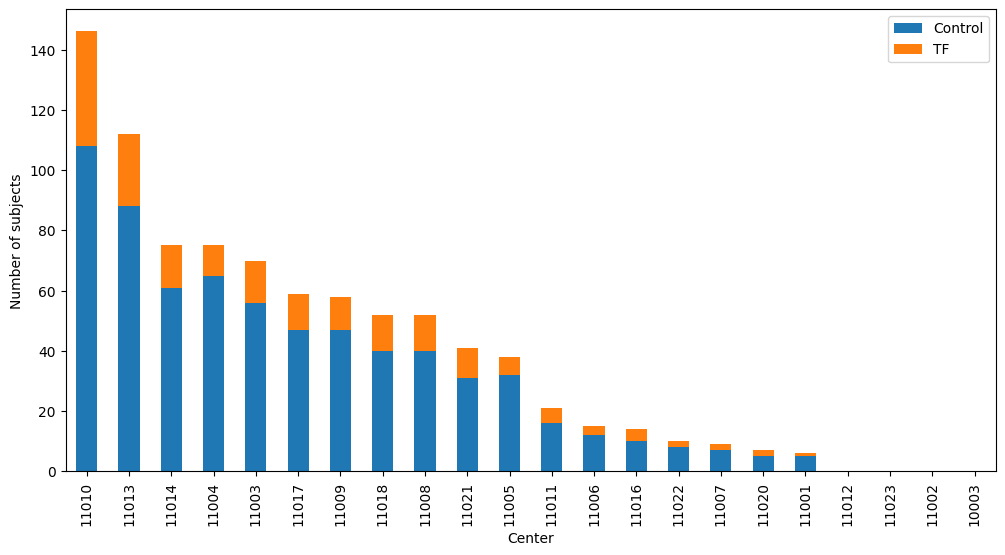}
    \caption{Number of patients that are selected as test set in idividual centers.}
    \label{fig:counts_testset}
\end{figure}

Fig.~\ref{fig:counts} provides a visual representation of the patient distribution across multiple centers in descending order of the number of patients. As anticipated, the distribution of subjects is not uniform; certain centers have an abundance of patients, with numbers exceeding a thousand, while others have fewer than ten patients. To ensure a fair and comprehensive evaluation of our model, we meticulously constructed a test set that accurately reflects the diversity and distribution of patients across these centers. This test set was created by randomly selecting 20\% of the entire cohort of clopidogrel patients, taking into account both the prevalence of treatment failures and the distribution of patients across the various centers. It's important to note that this test set was exclusively reserved for model evaluation and was deliberately excluded from any aspect of the training process. Fig.~\ref{fig:counts_testset} shows the distribution of selected patients in the test set.

\subsection{Federated learning}
To consolidate the model parameters from multiple participating centers, we used FedAvg algorithm implemented in NVIDIA NVFlare framework~\cite{rothNVIDIAFLAREFederated2023}. In the FedAvg algorithm, each participating center trains a local machine learning model using its own data. These local models are then communicated to a central server, which computes a weighted average of these models to create a global model update. This global model update is subsequently sent back to each participating device, where it is incorporated into the local model. This iterative process continues until the global model converges to a desirable state.

\begin{table}[t!]
\centering
\begin{tabular}{c|c}
    \hline
    Architecture & AUC \\ \hline \hline
    FCN &  0.793 \\ \hline
    GRU & 0.957 \\ \hline    
\end{tabular}
\caption{AUCs of ML architectures in centralized training scenario.}
\label{table:centralized_auc}
\end{table}

\begin{figure}[t!]
    \centering
    \includegraphics[width=1\linewidth]{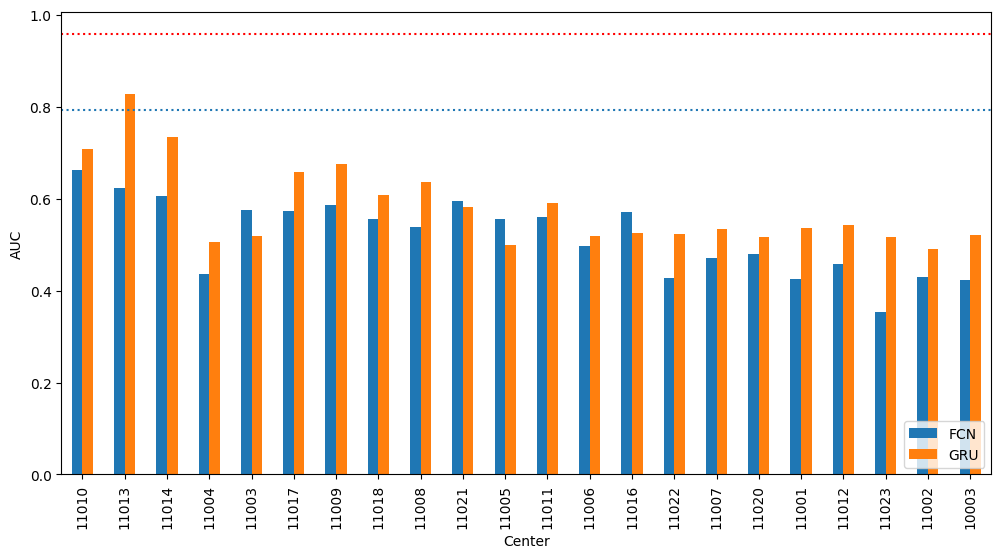}
    \caption{AUCs of locally trained models with their own data.}
    \label{fig:local_model}
\end{figure}
\begin{figure}[t!]
    \includegraphics[width=1\linewidth]{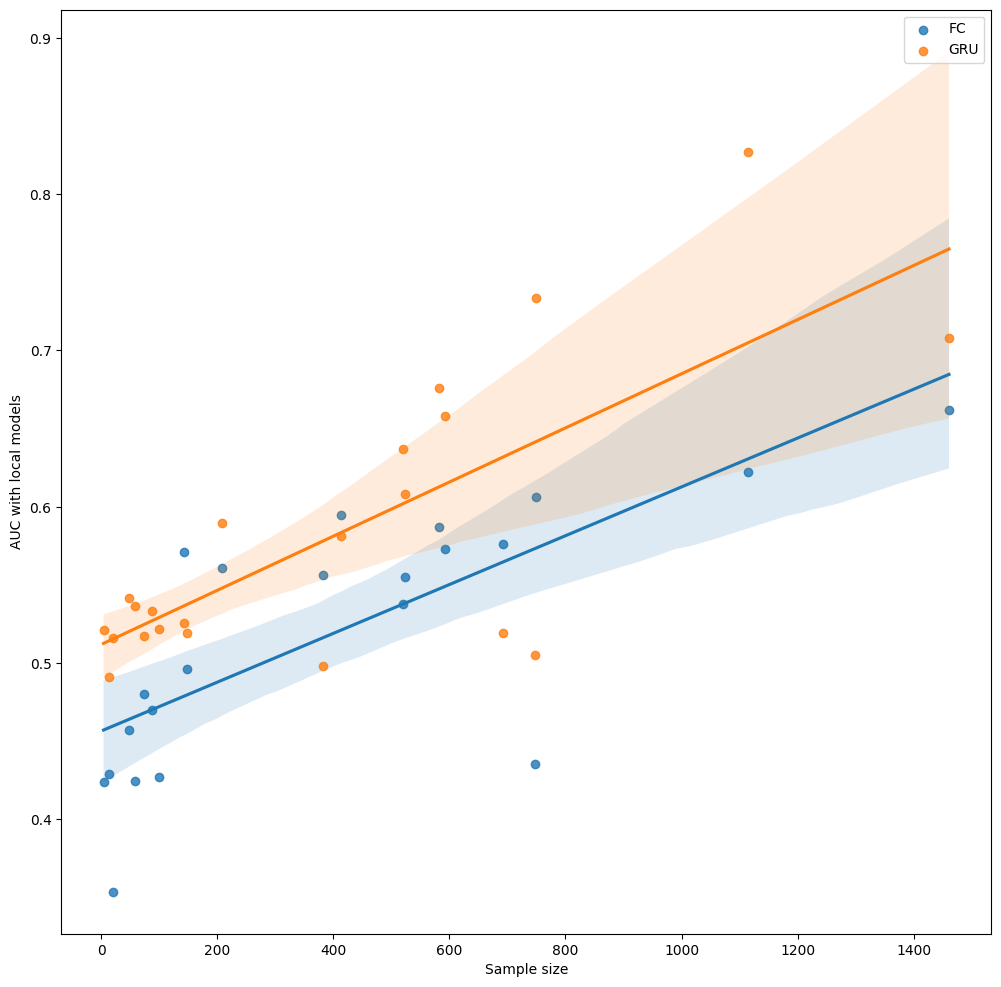}
    \caption{Scatter plot of the AUCs of locally trained models with respect to their cohort sizes.}
    \label{fig:size_vs_aug}
\end{figure}

\begin{figure*}[t!]
    \centering
    \begin{subfigure}[t]{0.49\linewidth}   
    \includegraphics[width=1\linewidth]{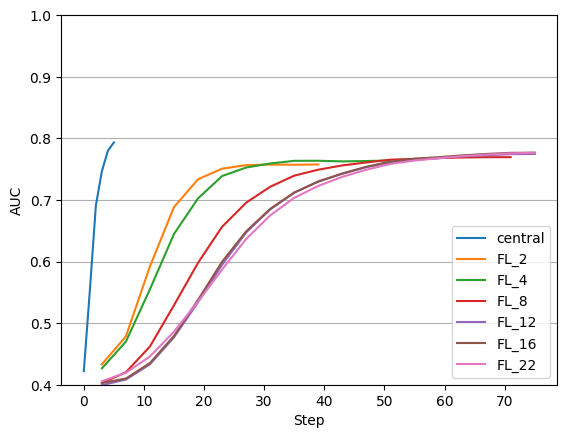}    
    \caption{FCN - Learning Curves}
    \end{subfigure}    
    \begin{subfigure}[t]{0.49\linewidth}   
    \includegraphics[width=1\linewidth]{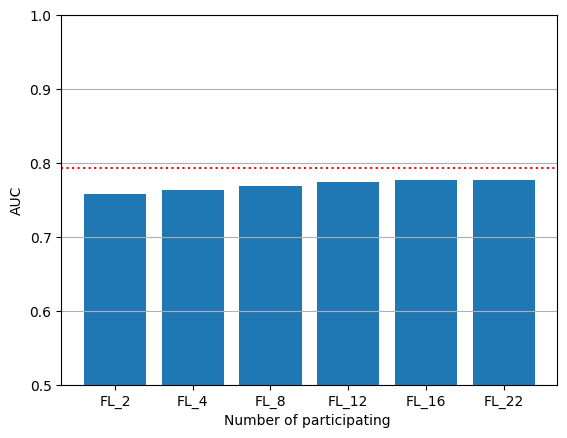}
    \caption{FCN - AUC}
    \end{subfigure}
    \begin{subfigure}[t]{0.49\linewidth}   
    \includegraphics[width=1\linewidth]{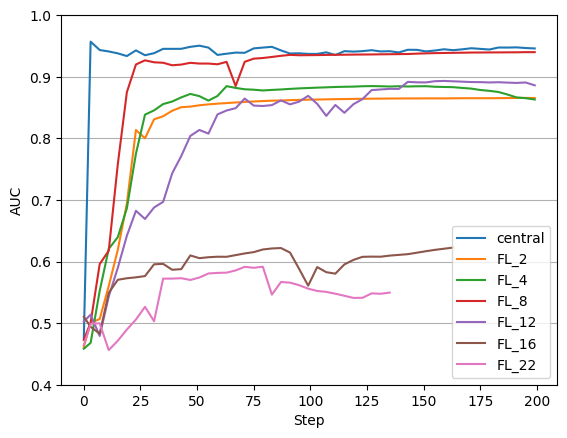}        
    \caption{GRU - Learning Curves}
    \end{subfigure}
    \begin{subfigure}[t]{0.49\linewidth}       
    \includegraphics[width=1\linewidth]{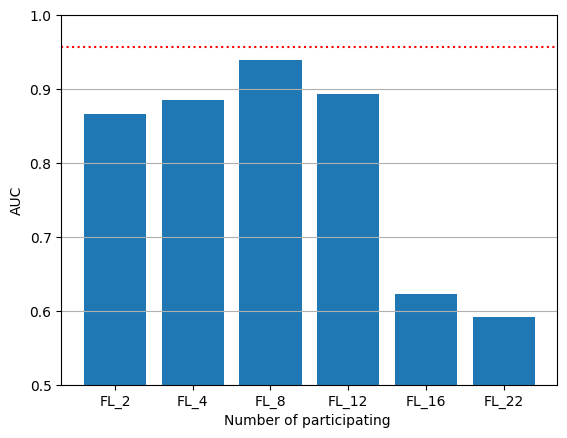}
    \caption{GRU - AUC}
    \end{subfigure}
    \caption{AUCs of the models that are trained in a federated learning scenario: (a)(b) for the FCN architecture and (c)(d) for the GRU architecture.}
    \label{fig:aucs}
\end{figure*}

\section{Experimental Results}

We primarily used two neural network architectures; fully connected network (FCN) and gated recurrent unit (GRU). While we applied a multi-hot encoder for FCN, we flattened the codes of the visits and concatenated them across all visits so that a patient has a sentence-like sequence of codes that can be fed into GRU. Table~\ref{table:centralized_auc} shows the AUC values of each ML architecture in the conventional centralized training scenario. The results shows that the recurrent architecture outperforms the fully connected architecture. This performance advantage can be attributed to the GRU's ability to capture and model the evolving dynamics and temporal dependencies inherent in patients' medical histories, a capability that the FCN, relying on a bag-of-words approach, does not possess.

Figure~\ref{fig:local_model} provides a clear depiction of the Area Under the Curve (AUC) values attained by individual models trained locally, using only their respective locally available datasets. As anticipated, the performance of these local models notably lags behind that achieved in the centralized training scenario, indicated by the dotted lines in the figure.

In Figure~\ref{fig:size_vs_aug}, we explore the intriguing relationship between the size of the local dataset and the corresponding model performance. This figure reveals a consistent trend across both model architectures: as the size of the local dataset increases, the performance of the local model also improves. This observation aligns with our expectations, as larger datasets enable models to discern more generalized patterns and enhance their predictive capabilities. This phenomenon underscores the significance of data quantity in the training process, emphasizing that access to more extensive and diverse datasets can yield models with superior generalization and prediction abilities, regardless of the chosen architecture.

In our federated learning scenario, we conducted a series of experiments involving varying numbers of participating centers. To simplify the analysis, we organized the centers in descending order based on the number of subjects within each center, incrementally incorporating them into the federated learning process.

Fig.~\ref{fig:aucs} illustrates the performance of federated learning concerning the number of participating centers. While it is evident that conventional centralized training achieves the highest Area Under the Curve (AUC) values and converges most rapidly when compared to models trained through federated learning, it is noteworthy that the performance gap between these approaches can be substantially narrowed under certain federated learning configurations.

Specifically, our FCN model achieved an AUC of 0.777 with the involvement of all 22 participating centers, while the GRU model reached an impressive AUC of 0.940 with the participation of just 8 centers. However, it's important to observe that the performance starts to degrade as the number of participating centers increases beyond a certain point, especially for the GRU architecture. This trend suggests that the simple averaging method employed to consolidate model parameters from multiple centers may not be the most effective solution, particularly when models from some participating centers exhibit poor performance, as observed in the case of the GRU architecture.

As part of our future research endeavors, we plan to delve deeper into the consolidation of model parameters, exploring methodologies that assign varying weights to these parameters based on their individual performance. This approach holds the potential to further enhance the performance of federated learning models in scenarios involving a diverse range of participating centers.

\section{Conclusion}

Our study highlighted the potential of federated learning in addressing clopidogrel treatment failure detection, offering a promising avenue for enhancing patient care through personalized treatment strategies while respecting data privacy. As we look to the future, our research paves the way for further investigations into advanced model consolidation techniques that can optimize federated learning in healthcare scenarios.

\bibliographystyle{IEEEbib}
{\bibliography{mybib}}

\end{document}